\documentclass{article}
\usepackage{spconf,amsmath,graphicx}
\usepackage{amssymb}
 \usepackage{url}

\usepackage{enumitem}
\setlist{nosep, leftmargin=14pt}
\usepackage{placeins}

\title{Improving clinical interpretability of linear neuroimaging models through feature whitening}

\name{
Sara Petiton$^1$,
Antoine Grigis$^1$,
Raphaël Vock$^1$,
Edouard Duchesnay$^1$
}
\address{$^1$University Paris-Saclay, CEA, CNRS, NeuroSpin, Baobab UMR 9027, Saclay, France}

\begin{document}
%
\maketitle
\begin{abstract}
Linear models are widely used in computational neuroimaging to identify biomarkers associated with brain pathologies. However, interpreting the learned weights remains challenging, as they do not always yield clinically meaningful insights. This difficulty arises in part from the inherent correlation between brain regions, which causes linear weights to reflect shared rather than region-specific contributions. In particular, some groups of regions, including homologous structures in the left and right hemispheres, are known to exhibit strong anatomical correlations. In this work, we leverage this prior neuroanatomical knowledge to introduce a whitening approach applied to groups of regions with known shared variance, designed to disentangle overlapping information across correlated brain measures. We additionally propose a regularized variant that allows controlled tuning of the degree of decorrelation. We evaluate this method using region-of-interest features in two psychiatric classification tasks, distinguishing individuals with bipolar disorder or schizophrenia from healthy controls. Importantly, unlike PCA or ICA which use whitening as a dimensionality reduction step, our approach decorrelates anatomically informed pairs of neuroanatomical regions while retaining the full input signal, making it specifically suited for feature interpretation rather than feature selection. Our findings demonstrate that whitening improves the interpretability of model weights while preserving predictive performance, providing a robust framework for linking linear model outputs to neurobiological mechanisms.
\end{abstract}
\section{Introduction}
\label{sec:intro}
Linear models are widely used in computational neuroimaging due to their simplicity, interpretability, and ability to uncover relationships between brain measurements and clinical outcomes. They are applied in a variety of contexts, including classification of psychiatric disorders, prediction of cognitive scores, and identification of biomarkers from brain imaging \cite{nunes_using_2020, nielsen_machine_2020}. However, interpreting linear weights can be challenging when features are correlated, a common scenario in neuroimaging, where brain regions or imaging-derived measures often covary \cite{haufe_interpretation_2014}. For example, model weights assigned to left and right hemispheric measurements of symmetrical regions, or to gray matter (GM) and cerebrospinal fluid (CSF) volumes within the same region, often capture shared variance rather than individual contributions, thereby limiting clinical interpretability. While “forward models” \cite{haufe_interpretation_2014} address collinearity \textit{post hoc}, they ignore known brain structure, such as inter-hemispheric correlations, which we explicitly incorporate during model fitting in the present work.

Whitening, or sphering, is an orthogonalization technique originally introduced in signal processing to remove linear dependencies between features \cite{de_cheveigne_joint_2014}. It has since been adopted across various domains to improve model stability and interpretability. In machine learning, whitening has recently gained attention in self-supervised learning (SSL), as a strategy to prevent feature collapse and enhance latent representations \cite{ermolov_whitening_2020}. Emerging studies also suggest its potential to enhance interpretability of convolutional neural networks \cite{chen_concept_2020}. In neuroimaging, whitening has primarily been applied to EEG or fMRI data for noise reduction \cite{olszowy_accurate_2019,engemann_automated_2015}. Recent work has begun evaluating whitening as a preprocessing step to improve the performance of post-hoc XAI methods in general image classification settings~\cite{clark_whitening_2026}, finding benefits that vary by method and architecture, but has no applications in neuroimaging contexts yet. Overall, whitening's potential for enhancing the interpretability of brain MRI models remains largely unexplored.

In this work, we leverage correlation-based zero-phase component analysis (ZCA-cor) whitening to enhance the interpretability of linear models trained on region-of-interest (ROI)-based brain features. ZCA-cor whitening has been identified as an optimal procedure for generating sphered variables that closely preserve the structure of the original data \cite{kessy_optimal_2018}.
We implemented both ZCA-cor, and an original regularized ZCA-cor as custom transformers compatible with the scikit-learn API.
Whitening is applied in two steps: (i) between left and right hemisphere measures of the same region, to disentangle hemisphere-specific contributions; and (ii) between gray matter (GM) and cerebrospinal fluid (CSF) volumes within each brain region, given their typically inverse relationship.

We evaluate the approach on two classification tasks, distinguishing individuals with bipolar disorder (BD) or schizophrenia (SCZ) from healthy controls (HC). Our results demonstrate that whitening and regularized whitening improve the interpretability of model weights without degrading predictive performance, offering a robust framework for linking linear model estimates to neurobiological mechanisms. 

\section{Materials and Methods}
\label{sec:format}

\subsection{Datasets}
\label{ssec:subhead}

Classification tasks for BD and SCZ were conducted using aggregated multi-site neuroimaging datasets. For BD, we used the BIOBD and BSNIP datasets \cite{biobd, bsnip}, comprising 861 participants across 12 acquisition sites. Among them, 56.4\% are female, 44.1\% were diagnosed with BD, and the mean age is $37.69 \pm 11.75$ years. For SCZ, classification was performed using the SCHIZCONNECT-VIP dataset \cite{schizconnect}, totaling 604 participants across 4 acquisition sites. 38.2\% of its participants are female, 45.5\% were diagnosed with SCZ, and the mean age is $33.33 \pm 12.34$ years.

Voxel-based morphometry (VBM) measures were extracted from T1-weighted brain MRI data using the CAT12 toolbox \cite{gaser_cat_2024}, (v 12.7). The brain was parcellated according to the Neuromorphometrics atlas \cite{neuromorphometrics}, which defines $140$ ROIs across both hemispheres. Both GM and CSF volumes were included, resulting in a total of $280$ ROIs.

\subsection{Machine Learning setup}
\label{ssec:subhead}

For both classification tasks (i.e., BD vs. HC, SCZ vs. HC), train and test sets were constructed using a ten-fold cross-validation (CV). Classification performance was assessed using the area under the receiver operating characteristic curve (ROC-AUC) and balanced accuracy (BAcc) metrics.
ROI-based measures were stratified by age, sex, acquisition site, and diagnosis using the MULM \cite{mulm} package. For whitening experiments, features were first residualized, then whitened, and finally scaled.
We chose logistic regression as our classification model for its interpretability and linearity, allowing for straightforward computation of weights in the whitened space and their projection back to the original feature space. The regression was performed with scikit-learn (v. 1.3.2) \cite{pedregosa_scikit-learn_2012}, with hyperparameter tuning via a grid search, to identify the best $L^2$ regularization parameter $C$, evaluated over the set $\{0.001, 0.01, 0, 1, 10\}$.

\subsection{Identification of highly covarying features}
\label{ssec:subhead}
We whitened features pairwise to corresponding left and right hemispheric regions (e.g., left amygdala, right amygdala), and to GM and CSF volumes of the same brain region (e.g., left amygdala GM, left amygdala CSF).
We selected such pairs based on prior evidence that corresponding regions in the left and right hemispheres exhibit strong correlations, and on the inverse relationship between GM and CSF volumes, such that higher GM volume corresponds to lower CSF volume within the same brain region. Although we demonstrate the approach on ROI-level features for simplicity, the framework naturally extends to voxel-wise data by whitening pairs of spatially symmetric voxels across hemispheres, or pairs of GM and CSF voxel intensities within the same region.

\subsection{Choice of whitening method}
\label{ssec:subhead}
Several whitening formulations exist, including principal component analysis (PCA) whitening, zero-phase component analysis (ZCA) whitening, and Cholesky whitening. These methods are reviewed in \cite{kessy_optimal_2018}, where ZCA-cor whitening is recommended as the ideal whitening procedure to generate sphered variables that remain as close as possible to the original features. ZCA-cor whitening differs from standard ZCA in that it uses the correlation matrix instead of the covariance matrix, operating on standardized features to produce decorrelated variables that remain maximally aligned with the original data and improving numerical stability. 

\subsection{ZCA-cor whitening}
\label{ssec:subhead}

Given $p$ a pair of correlated features (e.g., left/right amygdala GM volume, or GM/CSF of left hippocampus), let $X_{p} \in \mathbb{R}^{n \times 2}$ denote the sequence of observations of those features over $n$ subjects. Each feature is standardized to have zero mean and unit variance. We then compute the singular value decomposition (SVD) of the $2\times 2$ correlation matrix $R_{p}$ of the (standardized) $X_{p}$: $R_{p} = U_{p} \Lambda_{p} U_{p}^\top$,
where $U_{p}$ is the matrix of singular vectors of $R_{p}$ and $\Lambda_{p}$ is the corresponding diagonal matrix of singular values. We opt for SVD over eigenvalue decomposition for numerical stability.
We then define the whitening matrix $W_{p}$ by
\begin{equation}
W_{p} = U_{p} \Lambda_{p}^{-1/2} U_{p}^\top,
\end{equation}
and the whitened features by
\begin{equation}
Z_{p} = X_p W_p, \quad \text{with} \quad \mathrm{Corr}(Z_{p}) = I_2,
\end{equation}
A linear classifier is then trained on the whitened data: $\hat{y} = Z \beta + b$, where $Z$
 represents the whitened input features (i.e., the stacked matrices $Z_p$), $\beta$ denotes the classifier's weights in the whitened space, $b$ the bias term, and $\hat{y}$ the predicted scores.

Let $W_{\text{ZCA-cor}}$ denote the block-diagonal whitening matrix formed by combining the pairwise whitening matrices $W_p$. An interpretable classifier $\hat y = X\theta + b$ (whose inputs are the original features) is obtained by projecting the coefficients from whitened space to feature space, for each pair $p$, via the transformation:
\begin{align}\label{weight-remapping}
\theta = W_{\text{ZCA-cor}}^\top \beta.
\end{align}

\subsection{Mapping classifier weights back to feature space}
\label{ssec:weightinterp}

For each pair of whitened features $(z_{1}, z_{2})$ (each given as vector of length $n$), let $\beta_1$ and $\beta_2$ be the classifier weights associated with $z_{1}$ and $z_{2}$ in the whitened space, and $\theta_1$ and $\theta_2$ the corresponding weights in the original feature space.  

These weights are related by the $2\times 2$ whitening matrix
\[
W_{p} = \begin{bmatrix} w_{11} & w_{12} \\ w_{21} & w_{22} \end{bmatrix},
\]
where $w_{11} = w_{22} > w_{12} = w_{21}$ and $w_{11} > 0$, since a $2\times 2$ ZCA-cor whitening matrix is symmetric, with positive and equal diagonal terms larger than the off-diagonal entries.  

The original-space weights can be expressed as
\[
\theta_1 = w_{11}\beta_1 + w_{12}\beta_2, \quad 
\theta_2 = w_{21}\beta_1 + w_{22}\beta_2 = w_{12}\beta_1 + w_{11}\beta_2.
\]
It follows that
\[
\beta_1 > \beta_2 \hspace{.4em} \Rightarrow \hspace{.4em} (w_{11}-w_{12})(\beta_1-\beta_2) > 0 \hspace{.4em} \Rightarrow \hspace{.4em} \theta_1 > \theta_2.
\]
Thus, if $z_{1}$ has a higher weight than $z_{2}$ in the whitened space, the same order is preserved in the original feature space, preserving the relative contributions of input features (i.e., left/right or GM/CSF features).

\subsection{Regularized ZCA-cor whitening}
\label{ssec:alphazca}

Since our primary objective is interpretability, we propose a novel whitening procedure to decorrelate features while preserving class-relevant correlation structure. Partial whitening applied to pairs allows the classifier to retain inter-hemispheric relationships (e.g. some regions may exhibit stronger left-right correlations in patients than controls), which may reflect disease-related coupling patterns. 
Let $W_{\alpha}$ be the weighted whitening matrix:
\begin{equation}
W_{\alpha} = \alpha W + (1 - \alpha) I
\end{equation}
where $\alpha \in [0,1]$ controls the degree of decorrelation. When $\alpha=1$, the transformation reduces to the previously described ZCA-cor whitening.

\section{Results}
\label{sec:pagestyle}

\subsection{Feature disentanglement via whitening}
\label{ssec:subhead}
First, we applied a regularized ZCA-cor whitening to left-right pairs of homologous ROIs, setting $\alpha = 0.3$ to soften the whitening transformation and preserve class-specific covariance patterns. We then applied full ($\alpha = 1$) pairwise whitening to GM-CSF volumes within each region. These settings reflect that whitening matrices were estimated on the training set of each CV fold, including both patients and HC; while GM-CSF correlations are expected to be stable across individuals, left-right hemisphere correlations may vary across diagnostic classes. 

The ZCA-cor whitening matrices were computed using the training set and subsequently applied to the testing set for each of the 10 CV folds, thereby preventing data leakage. Figures 1 and 2 were obtained using the training set of the first fold of the BD dataset, using standardized features to accurately reflect the transformations applied before classification. In Figure 1, after whitening, the correlations between the homologous regions in the left and right hemispheres are markedly reduced, most clearly illustrated in the putamen (the four red squares at the center of the plot). Inverse correlations between GM and CSF are also attenuated, particularly visible in the bilateral amygdala (top-left of the plot). Figure 2 highlights the disentanglement of left/right and GM/CSF measures. After whitening, feature dispersion increases, with variance spread more evenly across hemispheres (bottom plot, hippocampus) and taking a more spherical distribution (top plot, anterior cingulate gyrus). Similar results were found with other folds and the SCZ dataset.

\begin{figure}[tb]
\centering
\includegraphics[width=7.3cm]{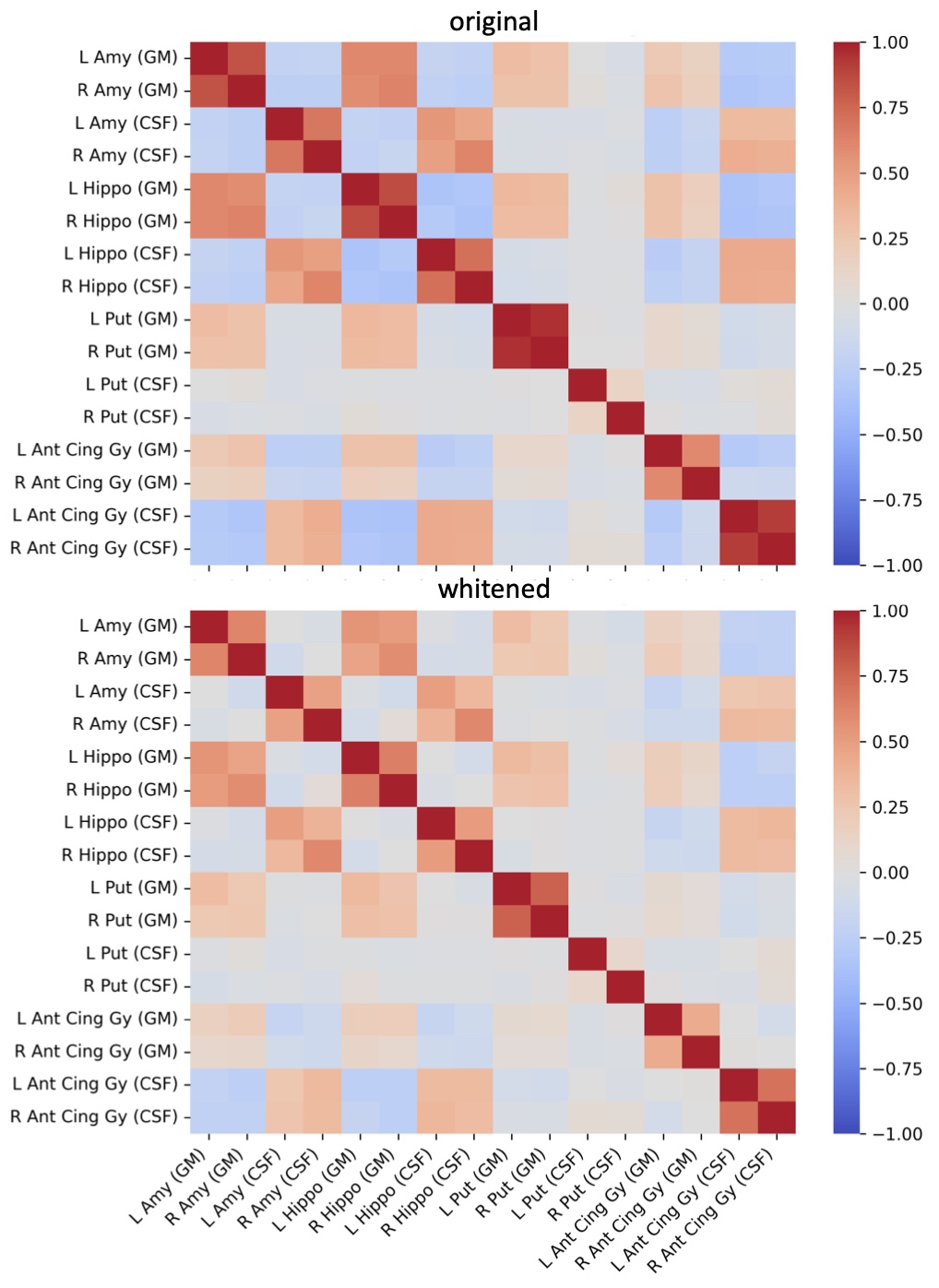}
\caption{Correlation matrices in feature (top) and whitened (bottom) space, computed on the BD dataset. Only amygdala, hippocampus, putamen, and anterior cingulate gyrus correlations were plotted for clarity and due to their known relationships to BD and SCZ.}
\label{fig:correlation}
\end{figure}
\FloatBarrier

\begin{figure}[tb]
\centering
\includegraphics[width=8 cm]{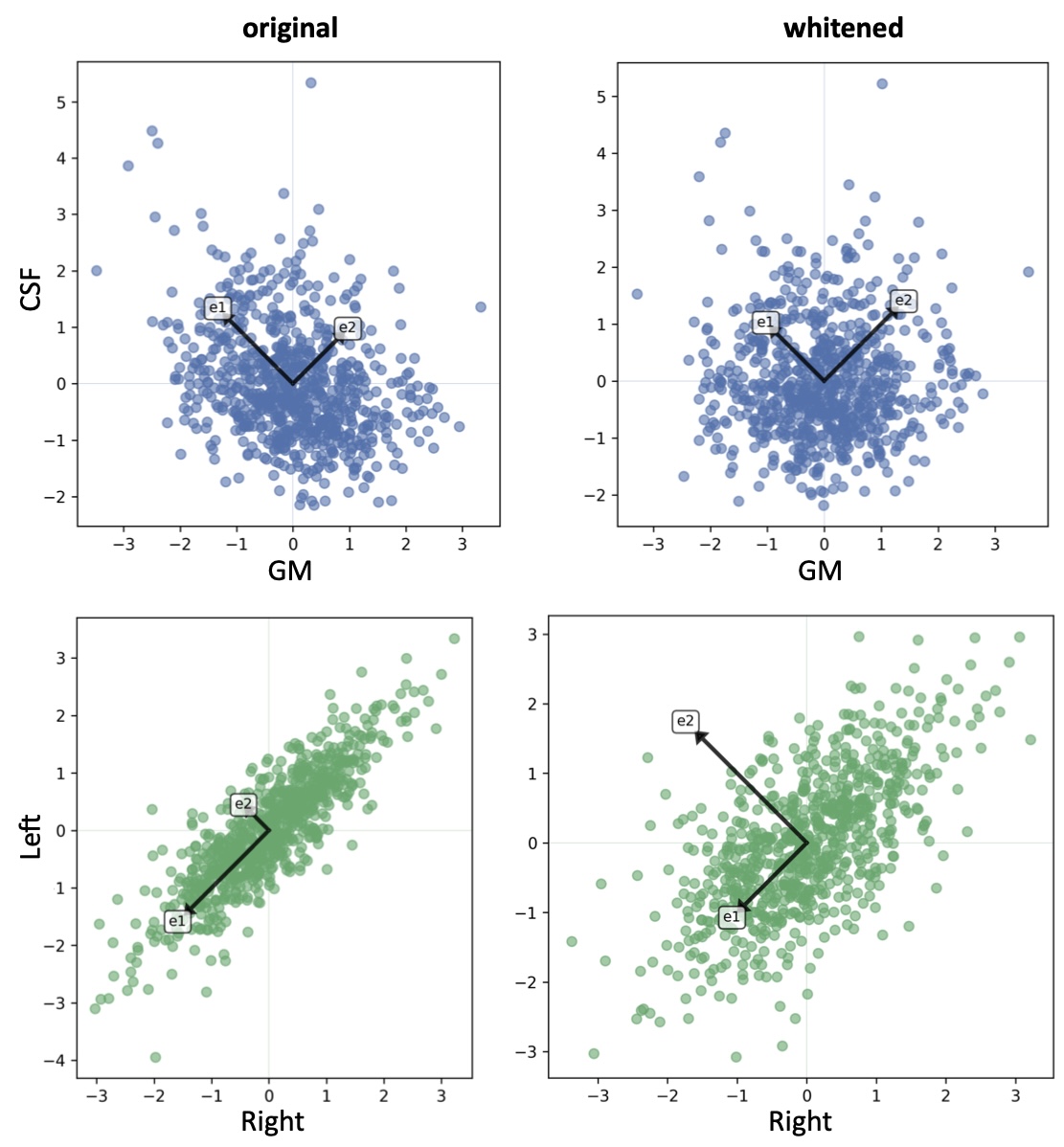}
\caption{Sphering of GM/CSF and left/right ROI pairs from the BD dataset. Top (blue): left anterior cingulate gyrus GM vs. CSF before (left) and after (right) whitening. Bottom (green): left vs. right hippocampus GM before and after whitening. Arrows show eigenvectors scaled up by 1.5 for clarity.}

\label{fig:whitening}
\end{figure}

\subsection{Whitening preserves classification performance}
\label{ssec:subhead}

Applying whitening did not affect classification performance (see Table~\ref{tab:results}). A Student's t-test comparing ROC-AUC and BAcc across cross-validation folds revealed no significant differences between classifiers trained on whitened versus unwhitened features ($p > 0.05$). In all experiments, the $L^2$ regularization parameter $C$ was fixed at $0.01$, as determined via grid search.

\begin{table}[tb]
\centering
\resizebox{\columnwidth}{!}{%
\begin{tabular}{lcccc}
\hline
\noalign{\vskip 5pt}
 & \multicolumn{2}{c}{\bf{BD vs. HC}} & \multicolumn{2}{c}{\bf{SCZ vs. HC}} \\
\cline{2-5}
\noalign{\vskip 5pt}
 & ROC-AUC & BAcc & ROC-AUC & BAcc \\
\hline
\noalign{\vskip 5pt}
\bf{Original} & $76.39 \pm 3.88$ & $69.58 \pm 4.15$ & $81.68 \pm 2.89$ & $73.76 \pm 4.27$ \\
\bf{Whitened}    & $76.24 \pm 4.37$ & $69.65 \pm 3.49$ & $81.0 \pm 3.12$ & $72.41 \pm 3.58$ \\
\noalign{\vskip 5pt}
\hline
\end{tabular}%
}
\caption{Classification test set results for BD vs. HC and SCZ vs. HC with and without whitening. Values are reported as percentages of mean $\pm$ standard deviation across CV folds.}
\label{tab:results}
\end{table}

\subsection{Whitening enhances interpretability}

Tables \ref{tab:top_features_stacked} and \ref{tab:top_features_sczi} report the mean coefficients estimated from logistic regression using whitened and unwhitened features for BD vs. HC and SCZ vs. HC classification, respectively. Weights computed on whitened features were mapped back to the original feature space as described in Equation~\ref{weight-remapping}. The listed regions correspond to the seven features with the largest absolute coefficients, reflecting their importance in classification.
Whitening improved the interpretability of regression coefficients by increasing their alignment with ENIGMA meta-analyses for both BD and SCZ (see \cite{cortical_bd, hibar_subcortical_2016, scz_cortical, scz_subcortical} for listings of ENIGMA-derived significant regions). In BD, whitening significantly strengthened correlations with cortical rankings and elevated subcortical regions (ventricles, hippocampus, thalamus, amygdala) closer to literature findings. In SCZ, whitening improved the ranking of key cortical and subcortical regions (hippocampus, inferior temporal gyrus, fusiform gyrus, lateral ventricles), even though overall correlations remained insignificant due to atlas differences. Across both disorders, whitening consistently enhanced the biological plausibility of classifier-derived ROI importance, underscoring its utility for more interpretable neuroimaging classification.

\begin{table}[tb]
\centering
\footnotesize
\resizebox{\columnwidth}{!}{%
\renewcommand{\arraystretch}{0.9} 
\begin{tabular}{lc}
\hline
\multicolumn{2}{c}{\textbf{Original}} \\
\hline
\textbf{Feature} & \textbf{Weight ($\pm$ std)} \\
\hline
R Pallidum (GM) & $0.196 \pm 0.010$ \\
R Cerebrum and Motor (GM) & $-0.138 \pm 0.007$ \\
L Pallidum (GM) & $0.136 \pm 0.014$ \\
R Superior Occipital Gyrus (GM) & $-0.120 \pm 0.017$ \\
L Parahippocampus Gyrus (CSF) & $-0.115 \pm 0.007$ \\
L Anterior Insula (GM) & $-0.111 \pm 0.006$ \\
L Medial Frontal Cerebrum (GM) & $0.111 \pm 0.011$ \\
\hline
\\[-1em]
\multicolumn{2}{c}{\textbf{Whitened}} \\
\hline
\textbf{Feature} & \textbf{Weight ($\pm$ std)} \\
\hline
R Pallidum (GM) & $0.249 \pm 0.013$ \\
R Fourth Ventricle (CSF) & $0.194 \pm 0.022$ \\
L Third Ventricle (CSF) & $0.176 \pm 0.060$ \\
L Middle Cingulate Gyrus (CSF) & $0.164 \pm 0.011$ \\
R Cerebrum and Motor (GM) & $-0.162 \pm 0.010$ \\
L Cerebral White Matter (CSF) & $0.140 \pm 0.012$ \\
L Parahippocampus Gyrus (CSF) & $-0.140 \pm 0.008$ \\
\hline
\end{tabular}
\renewcommand{\arraystretch}{1.0}
}
\caption{Top seven contributing brain regions and corresponding model weights (mean $\pm$ std) for BD vs. HC classification. The mean is computed over the 10 CV folds region-wise. }
\label{tab:top_features_stacked}
\end{table}
\FloatBarrier

\begin{table}[h!]
\centering
\resizebox{\columnwidth}{!}{%
\begin{tabular}{lc}
\hline
\multicolumn{2}{c}{\textbf{Original}} \\
\hline
\textbf{Feature} & \textbf{Weight ($\pm$ std)} \\
\hline
Right Pallidum (GM) & $0.261 \pm 0.012$ \\
Left Pallidum (GM) & $0.174 \pm 0.008$ \\
Right Putamen (GM) & $0.152 \pm 0.014$ \\
Left Putamen (GM) & $0.128 \pm 0.014$ \\
Left Accumbens (CSF) & $-0.116 \pm 0.007$ \\
Left Frontal Pole (GM) & $0.111 \pm 0.008$ \\
Right Angular Gyrus (CSF) & $-0.108 \pm 0.008$ \\
\hline
\\[-1em]
\multicolumn{2}{c}{\textbf{Whitened}} \\
\hline
\textbf{Feature} & \textbf{Weight ($\pm$ std)} \\
\hline
Right Pallidum (GM) & $0.311 \pm 0.015$ \\
Right Putamen (GM) & $0.169 \pm 0.019$ \\
Left Exterior Cerebellum (CSF) & $-0.164 \pm 0.012$ \\
Left Hippocampus (CSF) & $-0.159 \pm 0.013$ \\
Right Central Operculum (CSF) & $0.145 \pm 0.017$ \\
Right Cerebrum and Motor (CSF) & $-0.137 \pm 0.022$ \\
Left Anterior Cingulate Gyrus (CSF) & $-0.136 \pm 0.018$ \\
\hline
\end{tabular}%
}
\caption{Top seven contributing brain regions and corresponding model weights (mean $\pm$ std) for SCZ vs. HC classification. The mean is computed over the 10 CV folds region-wise.}
\label{tab:top_features_sczi}
\end{table}

\section{Conclusion}
\label{sec:typestyle}

In this study, we investigated the use of a novel regularized ZCA-cor whitening approach to enhance the interpretability of linear classifiers trained on ROI-based neuroimaging features. Pairwise whitening was performed between left and right hemispheres as well as between GM and CSF volumes, and classifier weights were subsequently projected back into the original feature space. Our results demonstrate that whitening preserves classification performance while improving the alignment of model-derived region importance as established by the ENIGMA meta-analytic results. While we apply our method to pairs of regions informed by prior knowledge of brain structure, applying it to the full set of features would yield results closely related to Haufe’s forward models \cite{haufe_interpretation_2014}.

While whitening has been applied in neuroimaging primarily as a dimensionality reduction technique (e.g., via PCA or ICA) or for noise reduction in EEG and fMRI signals \cite{olszowy_accurate_2019, engemann_automated_2015}, it had not previously been used to improve the interpretability of neuroimaging models. Looking ahead, while whitening extensions have been explored in deep learning for feature interpretation \cite{chen_concept_2020}, to our knowledge, this approach has not yet been adapted to neuroimaging applications. Our work addresses this gap and provides a foundation for extending the proposed methodology beyond linear models, integrating it into more complex architectures such as deep neural networks.

\section{Acknowledgments}
\label{sec:acknowledgments}
This research was generously supported by The Robert Debré Child Brain Institute (Paris) under grant ANR-23-IAHU-0010 and the research program in precision psychiatry (PEPR PROPSY, ANR-22-EXPR-0001), both of which are funded by the France 2030 program and the French National Research Agency (ANR). Additional support was provided by two ”Investissements d’Avenir” Hospital-University Research in Health initiatives: RHU-PsyCARE (ANR-18-RHUS 0014) and FAME (ANR-21-RHUS-0009).

\section{Compliance with ethical standards}
\label{sec:ethics}

This study was performed retrospectively using human participant data in accordance with local ethics guidelines.

\bibliographystyle{IEEEbib} 
\bibliography{whitening_bib}

\begin{thebibliography}{10}

\bibitem{nunes_using_2020}
Abraham Nunes et~al.,
\newblock ``Using structural mri to identify bipolar disorders – 13 site machine learning study in 3020 individuals from the enigma bipolar disorders working group,''
\newblock {\em Molecular Psychiatry}, vol. 25, no. 9, pp. 2130--2143, Sept. 2020.

\bibitem{nielsen_machine_2020}
Ashley~N. Nielsen et~al.,
\newblock ``Machine {Learning} {With} {Neuroimaging}: {Evaluating} {Its} {Applications} in {Psychiatry},''
\newblock {\em Biological Psychiatry: Cognitive Neuroscience and Neuroimaging}, vol. 5, no. 8, pp. 791--798, Aug. 2020.

\bibitem{haufe_interpretation_2014}
Stefan Haufe et~al.,
\newblock ``On the interpretation of weight vectors of linear models in multivariate neuroimaging,''
\newblock {\em NeuroImage}, vol. 87, pp. 96--110, Feb. 2014.

\bibitem{de_cheveigne_joint_2014}
Alain De~Cheveigné and Lucas~C. Parra,
\newblock ``Joint decorrelation, a versatile tool for multichannel data analysis,''
\newblock {\em NeuroImage}, vol. 98, pp. 487--505, Sept. 2014.

\bibitem{ermolov_whitening_2020}
Aleksandr Ermolov et~al.,
\newblock ``Whitening for self-supervised representation learning,'' 2020.

\bibitem{chen_concept_2020}
Zhi Chen, Yijie Bei, and Cynthia Rudin,
\newblock ``Concept whitening for interpretable image recognition,''
\newblock {\em Nature Machine Intelligence}, vol. 2, no. 12, pp. 772--782, Dec. 2020.

\bibitem{olszowy_accurate_2019}
Wiktor Olszowy, John Aston, Catarina Rua, and Guy~B. Williams,
\newblock ``Accurate autocorrelation modeling substantially improves fmri reliability,''
\newblock {\em Nature Communications}, vol. 10, no. 1, pp. 1220, Mar. 2019.

\bibitem{engemann_automated_2015}
Denis~A. Engemann and Alexandre Gramfort,
\newblock ``Automated model selection in covariance estimation and spatial whitening of meg and eeg signals,''
\newblock {\em NeuroImage}, vol. 108, pp. 328--342, Mar. 2015.

\bibitem{clark_whitening_2026}
Benedict Clark, Stoyan Karastoyanov, Rick Wilming, and Stefan Haufe,
\newblock ``The effect of whitening on explanation performance,''
\newblock {\em arXiv preprint arXiv:2602.09278}, 2026.

\bibitem{kessy_optimal_2018}
Agnan Kessy, Alex Lewin, and Korbinian Strimmer,
\newblock ``Optimal whitening and decorrelation,''
\newblock {\em The American Statistician}, vol. 72, no. 4, pp. 309--314, Oct. 2018.

\bibitem{biobd}
S~Sarrazin, A~Cachia, F~Hozer, et~al.,
\newblock ``Neurodevelopmental subtypes of bipolar disorder are related to cortical folding patterns: An international multicenter study,''
\newblock {\em Bipolar Disorders}, vol. 20, pp. 721--732, 2018.

\bibitem{bsnip}
Carol~A. Tamminga et~al.,
\newblock ``Bipolar and schizophrenia network for intermediate phenotypes: Outcomes across the psychosis continuum,''
\newblock {\em Schizophrenia Bulletin}, vol. 40, no. Suppl 2, pp. S131--S137, 2014.

\bibitem{schizconnect}
{SchizConnect},
\newblock ``Schizconnect: A multi-site schizophrenia neuroimaging database,'' \url{https://www.schizconnect.org/}.

\bibitem{gaser_cat_2024}
Christian Gaser et~al.,
\newblock ``Cat: a computational anatomy toolbox for the analysis of structural mri data,''
\newblock {\em GigaScience}, vol. 13, pp. giae049, Jan. 2024.

\bibitem{neuromorphometrics}
{Neuromorphometrics},
\newblock ``Neuromorphometrics: Brain atlas and parcellation resources,'' \url{https://www.neuromorphometrics.com/}, 2025,
\newblock Accessed: Nov 14, 2025.

\bibitem{mulm}
{pylearn-mulm},
\newblock ``mulm: massive univariate linear model,'' \url{https://www.neurospin.fr/pylearn-mulm/}, 2021.

\bibitem{pedregosa_scikit-learn_2012}
Fabian Pedregosa et~al.,
\newblock ``Scikit-learn: Machine learning in python,''
\newblock {\em Journal of Machine Learning Research}, 2012.

\bibitem{cortical_bd}
D~P Hibar et~al.,
\newblock ``Cortical abnormalities in bipolar disorder: an mri analysis of 6503 individuals from the enigma bipolar disorder working group,''
\newblock {\em Molecular Psychiatry}, vol. 23, no. 4, pp. 932--942, Apr. 2018.

\bibitem{hibar_subcortical_2016}
Hibar~D. P. et~al.,
\newblock ``Subcortical volumetric abnormalities in bipolar disorder,''
\newblock {\em Molecular Psychiatry}, vol. 21, no. 12, pp. 1710--1716, Dec. 2016.

\bibitem{scz_cortical}
TGM van Erp et~al.,
\newblock ``Cortical brain abnormalities in 4474 individuals with schizophrenia and 5098 control subjects via the enhancing neuro imaging genetics through meta analysis (enigma) consortium,''
\newblock {\em Biological Psychiatry}, vol. 84, no. 9, pp. 644--654, 2018.

\bibitem{scz_subcortical}
T~G~M Van~Erp et~al.,
\newblock ``Subcortical brain volume abnormalities in 2028 individuals with schizophrenia and 2540 healthy controls via the enigma consortium,''
\newblock {\em Molecular Psychiatry}, vol. 21, no. 4, pp. 547--553, Apr. 2016.

\end{thebibliography}

\end{document}